\documentclass{article}
\usepackage{arxiv}
\usepackage{natbib}
\usepackage{hyperref,cleveref}       
\usepackage{url}            
\usepackage{booktabs}       
\usepackage{amsfonts}       
\usepackage{nicefrac}       
\usepackage{microtype}      
\usepackage{lipsum}		
\usepackage{graphicx}
\usepackage{natbib}
\usepackage{doi}

\usepackage{enumitem}
\usepackage{hyperref}
\usepackage{url}
\usepackage{multirow}
\usepackage{graphicx}
\usepackage{booktabs}
\usepackage{svg}
\usepackage{color}
\definecolor{Gray}{rgb}{0.8,0.9,0.9}

\title{Integration of Radiomics and Tumor Biomarkers in Interpretable Machine Learning Models}

\author{Lennart Brocki and Neo Christopher Chung \\
Institute of Informatics, University of Warsaw, Poland }

\begin{document}
\maketitle
\begin{abstract}
Despite the unprecedented performance of deep neural networks (DNNs) in computer vision, their practical application in the diagnosis and prognosis of cancer using medical imaging has been limited. One of the critical challenges for integrating diagnostic DNNs into radiological and oncological applications is their lack of interpretability, preventing clinicians from understanding the model predictions. Therefore, we study and propose the integration of expert-derived radiomics and DNN-predicted biomarkers in interpretable classifiers which we call \emph{ConRad}, for computerized tomography (CT) scans of lung cancer. Importantly, the tumor biomarkers are predicted from a concept bottleneck model (CBM) such that once trained, our ConRad models do not require labor-intensive and time-consuming biomarkers. In our evaluation and practical application, the only input to ConRad is a segmented CT scan. The proposed model is compared to convolutional neural networks (CNNs) which act as a black box classifier. We further investigated and evaluated all combinations of radiomics, predicted biomarkers and CNN features in five different classifiers. We found the ConRad models using non-linear SVM and the logistic regression with the Lasso outperform others in five-fold cross-validation, although we highlight that interpretability of ConRad is its primary advantage. The Lasso is used for feature selection, which substantially reduces the number of non-zero weights while increasing the accuracy. Overall, the proposed \emph{ConRad} model combines CBM-derived biomarkers and radiomics features in an interpretable ML model which perform excellently for the lung nodule malignancy classification. 
\end{abstract}
\keywords{deep learning \and interpretability \and explainability \and concept bottleneck \and radiomics \and feature engineering }

Cancer kills about 10 million people annually, as a leading cause of death worldwide. Lung cancer is the most prevalent type of cancer in the world \citep{WHOcancer} and kills more patients than any other cancer in the United States \citep{CDCcancer}. Individuals suspecting or suffering from cancer routinely undergo medical imaging acquisition using computed tomography (CT), positron emission tomography (PET), and others, whose data (e.g., pixels and voxels) are becoming increasingly larger and more complex. Despite deep neural networks having shown unprecedented performance in computer vision, their application to medical images in clinical routine has been limited \cite{van2021deep,miotto2018deep}. One of the most important issues is a lack of interpretable machine learning models and a domain-specific implementation whose prediction can be understood by and communicated to clinicians \citep{rudin2022interpretable}. 

\begin{figure}
\centering
\includegraphics[width=.9\textwidth,keepaspectratio]{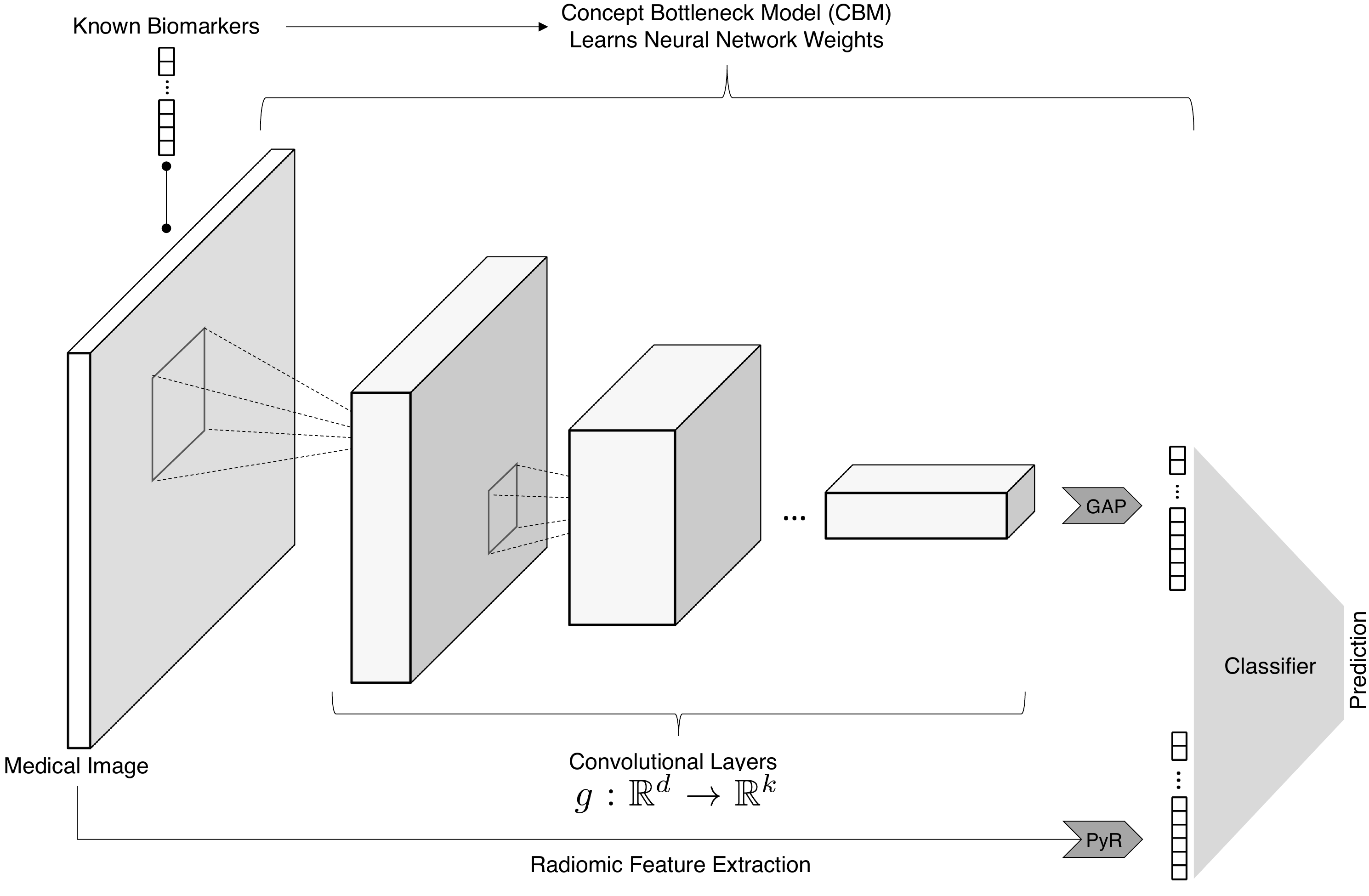}
\caption{The proposed \emph{ConRad} model integrates expert-derived radiomic features and biomarkers predicted from a concept bottleneck model (CBM) \citep{koh2020concept}. The final interpretable classifier uses both types of features competitively to classify a tumor as benign versus malignant.} 
\label{ConRad_architecture}
\end{figure}

To overcome this challenge in the context of lung nodule malignancy classification, we have developed and investigated an interpretable machine learning model called \emph{ConRad} which combines \textit{con}cept bottleneck models (CBM), which predict tumor biomarkers, and \textit{rad}iomic features, which are based on expert-derived characterization of medical images  (\cref{ConRad_architecture}). Both of these feature sets have the advantage that their meaning is clearly defined which enhances our model's transparency. Radiomics \citep{Lambin2017, YipAerts2016}. It has been shown to uncover cancer signatures that are visually indistinguishable for doctors and are able to identify cancer subtypes \citep{Hatt2019}. On the other hand, deep neural networks (DNNs) may result in high performance classifiers or features, that are not understood or explainable by oncological experts. As a proof-of-concept, we applied and evaluated different aspects of ConRad using the LIDC-IDRI (Lung Image Database Consortium and Image Database Resource Initiative) dataset \cite{armato2011lung}. 


Importantly, instead of using the annotated biomarkers directly in the malignancy prediction, we follow the concept bottleneck model (CBM) architecture \citep{koh2020concept} and train a DNN to predict the biomarkers which are then used in the final classifier. Therefore, once our model has been trained, biomarker annotations, which are expensive and time-consuming to obtain, are not needed anymore. Additionally, we perform standard radiomics feature extraction which provides interpretable statistical properties of cancer tumors. Both feature sets are fused to train the final classifier predicting benign vs. malignant tumors. Feature selection was explored by applying the Lasso (least absolute shrinkage and selection operator) \citep{lasso} to the logistic regression. The Lasso is shown to maintain, or even increase, the model performance while utilizing a substantially small number of features.

We have performed a comprehensive evaluation of various machine learning (ML) algorithms in ConRad models. Through 5-fold cross-validation, the performances are measured by the model accuracy, receiver operating characteristic (ROC) curves, and other metrics. On the other hand, we built end-to-end convolutional neural networks (CNNs), which directly predict the malignancy. From end-to-end CNNs, the CNN features are extracted and further used in independent ML classifiers for comparison. Both end-to-end CNNs and usage of CNN features represent uninterpretable black-box models, since radiologists do not gain any understanding about how the model is making its prediction. Our interpretable models perform comparable or superior to CNN.

There is an inherent trade-off between interpretability and accuracy in machine learning, including DNNs \citep{dziugaite2020tradeoff}. Instead of hyper-optimizing DNN architectures on LIDC-IDRI data, our study aims to engineer inherently interpretable features that are based on well-established radiological and oncological expertises. In the next \autoref{related}, we summarize related works, focusing on the development of radiomics and DNNs for oncology. Our ConRad model, data, and evaluation. are detailed in \autoref{methods}. Then \autoref{results} shows performance metrics of different ML classifiers and comparison against black-box classifiers based on CNNs. We also demonstrate informative feature selection, automatically achieved through use of the Lasso. Lastly, we summarize our findings and provide concluding remarks in \autoref{discussion}.

\section{Related Works}\label{related}
While the comprehensive review of DNNs applied on the lung nodule classification is beyond the scope of this paper, we note few related analysis of the LIDC-IDRI dataset \cite{armato2011lung}. Two of the earliest examples are \cite{shen2015multi} and \cite{kumar2015lung}, which use a supervised multi-scale approach and unsupervised feature extraction via an auto-encoder, respectively. Other approaches include 3D CNNs, which have been used by \citep{zhu2018deeplung}, so-called dilated CNN introduced in \citep{al2019gated} and curriculum learning \citep{al2022procan}, where the model is first trained on easy and later on harder samples, progressively growing the network in the process. All these approaches have in common that they focus on improving model accuracy and do not consider the interpretability of the developed models.

Our model shares similarities with the methods employed in \citep{causey2018highly,mehta2021lung,shen2019interpretable}, which all use features beyond the image pixels themselves. Radiomics features are combined with CNN features in \cite{causey2018highly}, but in contrast to ours, the annotated biomarkers are not used. The biomarkers are combined with CNN and radiomics features in \citep{mehta2021lung}, but the biomarkers are not predicted by a computational model and the annotations must be provided when the model is applied to unseen data. \cite{shen2019interpretable} uses DNNs to predict biomarkers, and extract features from previous layers of the biomarker predictor to be used in the malignancy classifier via a jump connection. Those features entering the final malignancy classifier are not interpretable by radiologists. Our evaluation of the best ML classifiers using multiple sets of features helped us to identify a high-performing classifier while maintaining greater interpretability.

\section{Methods}\label{methods}
\subsection{ConRad models and data}
The proposed \emph{ConRad} model is designed to extract different aspects of cancer images, leveraging well-established statistical properties and clinical variables of tumors. Particularly, our final model uses biomarkers predicted from a concept bottleneck model (CBM) \citep{koh2020concept} and 
radiomic features \citep{hancock2016lung} (visualized in \cref{ConRad_architecture}). CBM-predicted biomarkers and radiomic features are fed to ML classifiers to predict the tumor status (benign vs. malignant). Additionally, we built an end-to-end classifier based on a fine-tuned ResNet model \citep{he2016deep} which we use as a baseline comparison.

We particularly focus on CT images of lung tumors from the LIDC-IDRI (Lung Image Database Consortium and Image Database Resource Initiative) dataset \citep{armato2011lung}. The dataset consists of thoracic CT scans of 1018 cases alongside segmentations, the likelihood of malignancy, and annotated biomarkers for nodules with diameter $> 3$mm, obtained by up to four radiologists. The CT scans were processed using the \texttt{pylidc} package \citep{hancock2016lung}, which clusters the nodule annotations\footnote{This step is necessary in the presence of multiple nodules in a single scan since the dataset does not indicate which annotations belong to the same nodule. In case \texttt{pylidc} assigns more than four annotations to a nodule, the concerned nodule is not admissible.} and provides a consensus consolidation of the annotated nodule contours. The likelihood of malignancy is ranked from one (highly unlikely) to five (highly suspicious) and we aggregate the radiologists' different annotations by calculating the median, where nodules with a median of three are discarded as ambiguous and above or below three as benign or malignant. This procedure yields a total of 854 nodules, 442 of which are benign. 

Input samples were created by first isotropically resampling the CT scans to 1mm spacing and then extracting $32\times 32$ crops around the nodule center in the axial, coronal, and sagittal planes. Hounsfield unit values below -1000 and above 400 were clamped to filter out air and bone regions.

\subsection{Feature engineering and Model training}

First, we conduct the standard radiomic feature extraction from LIDC-IDRI samples using the \texttt{PyRadiomics} package \citep{PyRadiomics}. Particularly, three classes of radiomic features are calculated based on first-order statistics (18 features), 3D shapes (14 features), and higher-order statistics (75 features). All radiomic features are extracted from images where the tumor has been delineated using the segmentation masks provided by the LIDC data. First-order statistics include energy, entropy, centrality, and other distributional values are calculated. Shape features include volumes, areas, sphericity, compactness, elongation, and other descriptors of masked tumors. Higher order statistics are given by the Gray Level Co-occurrence Matrix, Gray Level Size Zone, and others.

Second, a ResNet-50 \citep{he2016deep} model pre-trained on the ImageNet \citep{deng2009imagenet} is fine-tuned to predict eight well-known clinical variables informative of tumors, in a concept bottleneck model (CBM) \citep{koh2020concept}. Instead of predicting the tumor status directly (benign vs malignant), the CBM associates the input samples with annotated biomarkers provided by the LIDC-IDRI dataset;  namely, subtlety, calcification, sphericity, margin, lobulation, spiculation, texture, and diameter. The values used for training the CBM are obtained by averaging the annotations provided by the different radiologists. 

Note, that the degree to which the described features are interpretable as well as in which way they are interpretable varies. Radiomics features are all interpretable in the sense that they follow from a rather simple mathematical definition. Since many of them represent rather abstract statistical features, their meaning may nonetheless be not intuitive for clinicians. The predicted biomarkers, on the other hand, are computed by an opaque algorithm (the DNN) but their meaning can be immediately understood. 

To match the input dimensions of the ResNet-50 model, input samples are upscaled to $224\times224$ and duplicated in each color channel. Input samples are z-normalized with mean and standard deviation obtained from the training set and identical normalization applied to the test set. We employ 5-fold cross-validation. The model is trained for 50 epochs using the Adam optimizer with PyTorch default parameters and a mini-batch size of 32. The initial learning rate is set to $10^{-3}$ and is annealed by multiplying with $0.1$ after 20 and 40 epochs. During training, an input sample is obtained by randomly selecting one of the three views and during testing, the model output is averaged across all three views to obtain the final output. The trained CBM is used to obtain predicted values for selected biomarkers from samples in the testing set. Unless specified, our evaluation is based on predicted biomarkers (instead of annotated values) to reflect clinical practices where labor-intensive manual segmentation and quantification may be unavailable. Note that we included clinical labels from LIDC-IDRI but depending on the data and domain experts, other biomarkers may be used to train the CBM.

Third, we have also extracted features from training a convolutional neural network (CNN) to directly predict the tumor status. For fair comparison, the same ResNet-50 architecture used in CBM was used to build CNNs. The preprocessing and training procedure is identical to the CBM described above. This end-to-end classifier gives us a baseline to be compared with our proposed interpretable ConRad model. Furthermore, we extract 512 features from the global average pooling layer of the fine-tuned ResNet-50, which are also fused with the radiomics features and CBM-derived biomarkers for downstream classification. We investigate whether a black box DNN model may have superior performance for classifying lung cancer images. And if so, what trade-off between interpretability and accuracy may exist and be acceptable for radiological applications.

\subsection{Machine Learning Classifiers}

There are three types of features available for training a model to classify the malignancy of nodules. Using each type of features and their combinations, we constructed a total of 7 datasets for the final classifer:
\setlist{nolistsep}
\begin{itemize}[noitemsep]
\itemsep0em 
  \item Biomarkers + Radiomics (ConRad models)
  \item Radiomics features
  \item Biomarkers (predicted by CBM)
  \item CNN features
  \item CNN + Radiomics
  \item CNN + Biomarkers
  \item CNN + Radiomics + Biomarkers (All)
\end{itemize}
Note that a nodule diameter appears in both Biomarkers and Radiomics feature sets. Having two features that are closely related would cause instability. Thus when necessary, we removed a diameter from Biomarkers. All datasets are z-score normalized on the training set and identical normalization is applied on the test set. 

As final-layer classifiers (\cref{ConRad_architecture}), we applied and evaluated linear and non-linear support vector machine (SVM) \citep{cortes1995svm}, logistic regression \citep{cox1958regression} with and without Lasso regularization \citep{lasso}, and random forest \citep{ho1995random}. The SVM and Lasso regularization parameters $C$ and $\lambda$, respectively, are selected via 5-fold cross-validation on the whole dataset. The Lasso (least absolute shrinkage and selection operator) adds $L_1$ penalty to the loss function, which may achieve feature selection with coefficients of 0. The Lasso has been applied on a large number of radiomic features to remove highly correlated features, such that the optimal model performance is achived with much less features \citep{koh2022aicancerreview}. For each combination of features, we track the number of features selected via the Lasso.

\begin{table}[t!]
    \centering
    \caption{Evaluation of different classifiers in ConRad models. Performance metrics are averaged over the five-fold cross-validation.}
    \begin{tabular}{llll}
    \toprule
        Final Layer Classifier & Recall & Precision & Accuracy \\ \midrule
        Non-linear SVM & 0.886 & \textbf{0.899} & \textbf{0.897} \\ 
        Linear SVM & 0.886 & 0.893 & 0.893 \\ 
        Random Forest & 0.879 & 0.883 & 0.881 \\ 
        Logistic Regression & 0.884 & 0.893 & 0.892 \\ 
        Logistic Regression with the Lasso & \textbf{0.896} & 0.893 & 0.896 \\ \bottomrule
    \end{tabular}
    \label{Conrad-Eval}
\end{table}

Overall, we built five classifiers using seven aforementioned datasets, combing three types of features. We applied an independent 5-fold cross-validation, where each classifier is trained on four folds and performance is evaluated on the held-out fold. The performance metrics (namely accuracy, precision, and recall) are calculated and averaged across 5-fold cross-validation. A receiver operating characteristic (ROC) curve is constructed by measuring false positive rates (FPR) and true positive rates (TPR) at a wide range of thresholds.  

Our proposed model can be contrasted with popular end-to-end DNN algorithms which are considered to be uninterpretable ``black box'' models. Instead, we are interested in creating an interpretable method, where the final prediction is explained by statistically and clinically relevant features. Furthermore, we leverage and expose oncological knowledge by directly utilizing radiomic features. Importantly, ML classifiers competitively use multiple types of features (radiomics and biomarkers) which helps to highlight their interplay. 

\section{Results}\label{results}

\subsection{Evaluation of ConRad models}

\begin{figure}[hbt]
\centering
\includegraphics[width=0.6\textwidth]{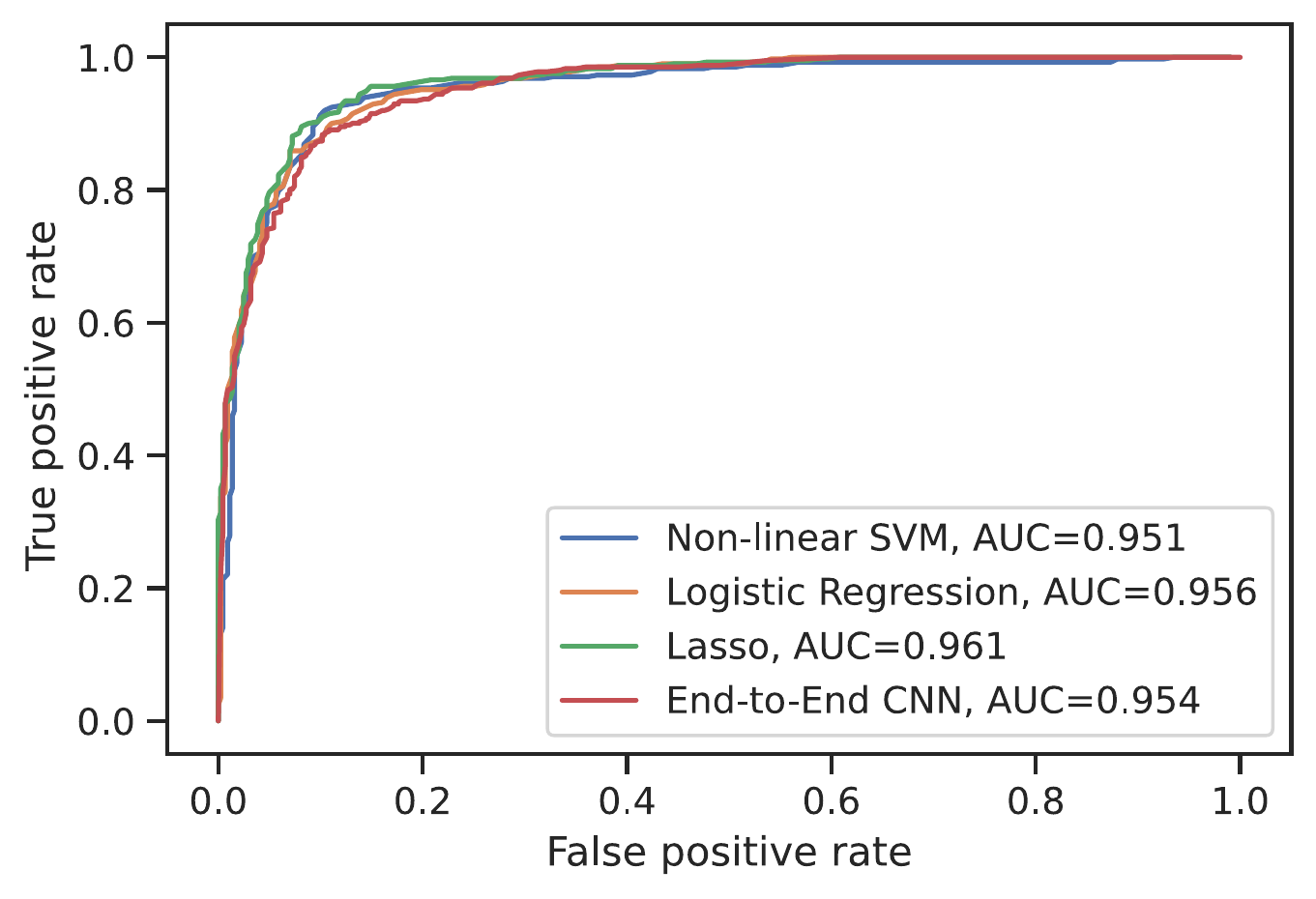}
\caption{ROC curves for ConRad models in comparison with the end-to-end CNN model. At varying thresholds, false positive and true positive rates are measured for each classifier, followed by averaging over 5 test sets. The areas under the ROC curves (AUCs) are similar in all considered classifiers, where the logistic regression with the Lasso outperforms.} 
\label{fig:roc}
\end{figure}

\emph{ConRad} models combine tumor biomarkers predicted from DNNs and radiomic features defined by radiological experts. LIDC-IDRI contains 8 biomarkers, which are used to construct a concept bottleneck model (CBM) using the pre-trained ResNet-50. We ran each of the CT scan through {\tt PyRadiomics} to obtain  first-order statistics (18 features), 3D shapes (14 features), and higher-
order statistics (75 features). We have trained and evaluated five different ML classifiers, based on a total of 115 features. 

The model accuracy, recall, and precision are averaged over five-fold cross-validation, where the test set was not used in training. \cref{Conrad-Eval} show that all the classifiers are reporting an accuracy in the range of $0.881-0.897$; non-linear (radial) SVM outperforms all other classifiers, whereas the Random Forest is the worst. For the majority of feature sets, linear SVM performs worse than a non-linear SVM. In terms of the ROC, all the ConRad models and the end-to-end CNN model perform similarly (\cref{fig:roc}).

When the Lasso (least absolute shrinkage and selection operator) was additionally applied on logistic regression, only 12 out of 114 features (see  ``Biomarkers+Radiomics'' in \cref{fig:lasso}) were selected. The logistic regression with the lasso increased the model performance compared to that without the lasso. Additionally, using a small set of features is inherently more interpretable, as 12 features can be readily visualized and understood. 

\subsection{Comparison to CNN models}

\begin{table}[!ht]
    \centering
    \caption{Model accuracy of baseline CNN models. Except the first row, all classifiers use CNN features. Compare to \cref{Conrad-Eval}. }
    \begin{tabular}{ll}
    \toprule
        Classifier & Accuracy \\ \midrule
End-to-end CNN classifer & \textbf{0.891} \\ 
        Non-linear SVM & 0.875 \\ 
        Linear SVM & 0.87 \\ 
        Random Forest & 0.888 \\ 
        Logistic Regression & 0.858 \\ 
        Logistic Regression with the Lasso & \textbf{0.891} \\ \bottomrule
    \end{tabular}
    \label{Baseline-CNN}
\end{table}

In order to compare the ConRad models to baselines, we constructed models that exclusively rely on CNN features, extracted from the end-to-end model outlined in \cref{methods}. The end-to-end CNN classifier itself yields $0.891$ accuracy and the other considered classifiers lie in the range from 0.858 to 0.891 (\cref{{Baseline-CNN}}). CNN-based classifiers did, therefore, not perform as well as the ConRad models (compare against \cref{Conrad-Eval}). What's more, radiomics and biomarkers are interpretable representing morphological or biological characteristics, whereas CNN features are not interpretable. Note that ConRad does not aim to maximize the model accuracy, and even slightly lower accuracy may be acceptable in exchange for higher interpretability.

However, there is still a concern that uninterpretable CNN features may contain additional information that could increase the ConRad models' performance. To test this idea, CNN features are added to ConRad models, creating the fullest models (``CNN + Radiomics + Biomarkers''). The five considered ML classifiers are trained on this feature set and performances are measured over five-fold cross-validation (complete results available in Appendix \cref{full_evaluation}). For all of the classifiers except random forest, adding CNN features on top of the radiomics and biomarkers decreases the accuracy. Table \ref{full_evaluation} shows that non-linear SVM with predicted biomarkers and radiomics (i.e., ConRad models) slightly outperforms other approaches in terms of precision and accuracy. We also observe that the inclusion of the Lasso improves logistic regression for most feature sets across recall, precision, and accuracy measures.


\subsection{The Lasso and Feature Selection}

\begin{figure}[t]
\centering
\includegraphics[width=0.9\textwidth]{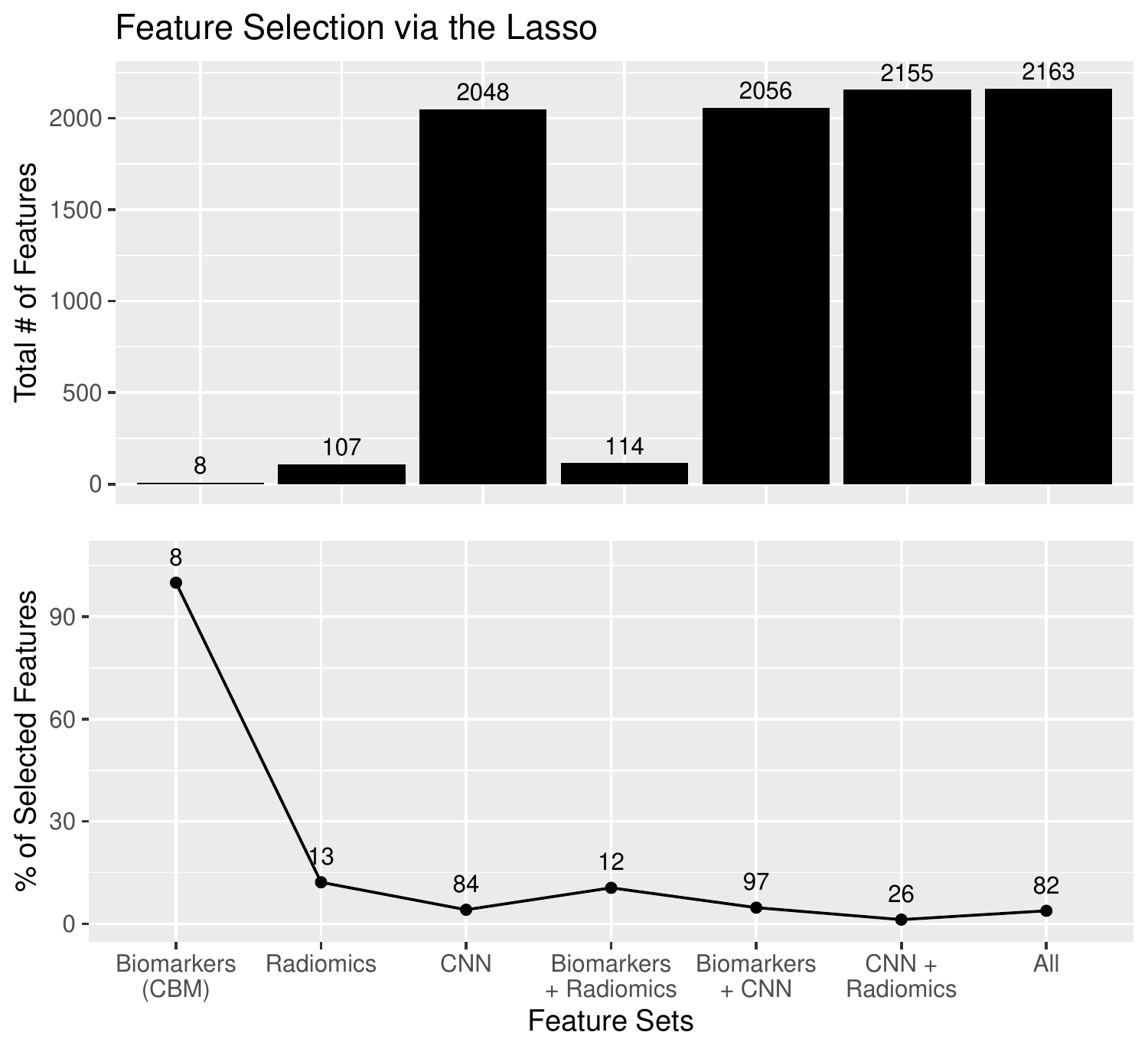}
\caption{Feature selection by the Lasso in logistic regressions. In each feature set, the Lasso ($L_{1}$ regularization) is applied with the penalty parameter selected via cross-validation. Except for biomarkers which only contain 8 features, only small percentages of features are selected.} 
\label{fig:lasso}
\end{figure}

In a regression model, the Lasso \citep{lasso} adds a $L_1$ penalty in the loss function which may reduce weights (i.e., coefficients in logistic regression), even to zero. The zero weights imply feature selection, where only a subset of all available features is used in classification. We found that the Lasso drastically reduces the number of considered features in most of the feature sets \cref{fig:lasso}. All 8 biomarkers are automatically selected by the Lasso, presumably due to their low dimension. When considering 2048 CNN features, only $4.10\%$ (84 features) were selected \cref{fig:lasso}.

Logistic regression with the Lasso does not only strongly reduce the number of selected features, but performs comparably or superiorly to the non-regularized logistic regression on almost every considered dataset and performance metric (\cref{full_evaluation}). On the pure radiomics features, the Lasso selects five shape features, seven higher-order statistics features, and the first-order minimum. With just $12.1\%$ of radiomic features, the accuracy remains the same. When combining biomarkers and radiomics features in the proposed ConRad approach, the Lasso selects five biomarkers and seven radiomics features. Using this subset of features ($10.5\%$ of features), the regularized logistic regression achieves $0.896$ accuracy, compared to the unregularized version with $0.892$ accuracy. Feature selection via the Lasso increases interpretability without losing model performance. 

\section{Discussion}\label{discussion}
We investigated different models to classify the lung nodule malignancy, with a focus on interpretability and explainability. Particularly, as feature engineering, a concept bottleneck model (CBM) predicts biomarkers. When applied, the proposed ConRad model does not need annotated biomarkers as often is a case in real world. In total, we investigated 35 classifiers, utilizing seven combinations of features and five ML algorithms. Additionally, an end-to-end CNN classifier was evaluated as a baseline. ConRad classifiers based on biomarkers and radiomics features show great performance on par with the end-to-end CNN model, which inherently act as a black box. Non-linear SVM and logistic regression with the Lasso showed the best performance generally.

The intermediate features that feed into the final classifier are interpretable and clinically meaningful by design. The biomarkers represent properties of nodules that have an intuitive meaning for clinicians, who can therefore gain understanding of the decision-making process of the model. This can nurture trust in the ML model since the users don't have to blindly believe the model's classifications. Instead, the users can comprehend, and even interact with, how the model reaches its decisions. For example, if a truly small nodule has been incorrectly predicted to have a large diameter, the clinician may recognize by visually inspecting the CT scan (i.e., a routine tasks). This may allow them to making informed decisions with a support of ML. Furthermore, in the ConRad model, the diameter prediction can be corrected by the clinician, directly resulting in updated prediction. Our ConRad model, therefore, lends itself to a human-in-the-loop (HITL) approach, that may help gaining deeper understanding and trust of the model operating characteristics.

Besides using independent training steps to create an interpretable classifier, we also consider feature selection in malignancy classification. When combining biomarkers and radiomics features, the number of selected features were drastically reduced with the Lasso \citep{lasso}. Many biomarkers and radiomics features may be correlated, thus removing redundancy can stabilize the model and improve the overall performance with translational potentials \citep{koh2022aicancerreview}. When applied to CNN features, extreme reduction to only $1-5\%$ features actually improved the model accuracy. Not only are CNN features unexplainable, a majority of them contain redundant information and do not contribute to the malignancy classification. In the future, this may be a fruitful way to investigate how to identify predictive features, while ensuring interpretability inherent in a manageable number of features.

The combination of biomarkers and radiomics features outperforms all other considered combinations, although the biomarkers demonstrated the most predictive power for the LIDC-IDRI data \citep{armato2011lung}. Some of the most critical information about lung nodules are contained in both feature sets, as exemplified by a diameter. Thus, we consider both feature sets to be critical part of bringing clinical knowledge into interpretable ML, especially as real world medical images may not contain as rich annotations as LIDC-IDRI. Pre-defined mathematical characterization of medical images in radiomics is always available and clinically validated \citep{Lambin2017, YipAerts2016}. We plan to investigate deeper into the interplay between these features, especially as we plan to apply ConRad to other medical imaging datasets.

Overall, the proposed ConRad model combines concept bottleneck models and radiomics to create an interpretable model, compared to end-to-end DNN classifiers. Transparency and explainability in a model prediction helps radiologists and oncologists to understand and form informed diagnosis and prognosis. Without that critical interpretability component, a black box classifier like end-to-end DNNs may harbor critical failure modes that is unknown and unknowable. Therefore, instead of focusing solely on model performance for medical applications, more investigations that consider interpretability are warranted to see broader incorporation of explainable AI in radiology and oncology.

\vspace{6pt} 


\section*{Acknowledgement}
This work was funded by the ERA-Net CHIST-ERA grant [CHIST-ERA-19-XAI-007] long term challenges in ICT project INFORM (ID: 93603), by the National Science Centre (NCN) of Poland [2020/02/Y/ST6/00071]. This research was carried out with the support of the Interdisciplinary Centre for Mathematical and Computational Modelling University of Warsaw (ICM UW) under computational allocation no GDM-3540; the NVIDIA Corporation’s GPU grant; and the Google Cloud Research Innovators program.

\bibliographystyle{unsrtnat}
\bibliography{ConRad}

\clearpage
\section*{Appendix}
\setcounter{table}{0}
\renewcommand{\thetable}{A\arabic{table}}

\begin{table}[h!]
\centering
\caption{Full comparison of five classifiers on 7 different combinations of features. Performance metrics are averaged over the five-fold cross-validation.}
\begin{tabular}{llccc} 
\toprule
Classifier & Features & Recall & Precision & Accuracy \\
\midrule
\multirow{5}{8em}{Non-linear SVM}&cnn & $0.869$ & $0.880$ & $0.875$ \\
&radiomics & $0.869$ & $0.873$ & $0.876$ \\
&biomarkers & $\mathbf{0.900}$ & $0.876$ & $0.890$ \\
&all & $0.869$ & $0.886$ & $0.879$ \\
&cnn+rad & $0.869$ & $0.878$ & $0.876$ \\
&bio+rad & $0.886$ & $\mathbf{0.899}$ & $\mathbf{0.897}$ \\
&bio+cnn & $0.876$ & $0.868$ & $0.891$\\
\midrule 
\multirow{5}{8em}{Linear SVM} & cnn & $0.859$ & $0.862$ & $0.870$ \\
&radiomics & $0.879$ & $0.880$ & $0.883$ \\
&biomarkers & $0.891$ & $0.893$ & $0.896$ \\
&all & $0.855$ & $0.870$ & $0.863$ \\
&cnn+rad & $0.864$ & $0.873$ & $0.868$ \\
&bio+rad & $0.886$ & $0.893$ & $0.893$ \\
&bio+cnn & $0.852$ & $0.861$ & $0.869$\\
\midrule 
\multirow{5}{8em}{Random forest} & cnn & $0.871$ & $0.887$ & $0.888$ \\
&radiomics & $0.876$ & $0.878$ & $0.878$ \\
&biomarkers & $0.891$ & $0.879$ & $0.889$ \\
&all & $0.881$ & $0.897$ & $0.885$ \\
&cnn+rad & $0.874$ & $0.894$ & $0.890$ \\
&bio+rad & $0.879$ & $0.883$ & $0.881$ \\
&bio+cnn & $0.891$ & $0.889$ & $0.890$\\
\midrule
\multirow{5}{8em}{Logistic regression} & cnn & $0.857$ & $0.858$ & $0.858$ \\
&radiomics & $0.874$ & $0.884$ & $0.883$ \\
&biomarkers & $0.886$ & $0.892$ & $0.893$ \\
&all & $0.871$ & $0.878$ & $0.875$ \\
&cnn+rad & $0.864$ & $0.864$ & $0.874$ \\
&bio+rad & $0.884$ & $0.893$ & $0.892$ \\
&bio+cnn & $0.850$ & $0.867$ & $0.874$\\
\midrule
\multirow{5}{8em}{Logistic regression with the Lasso (feature selection)} & cnn & $0.871$ & $0.876$ & $0.891$ \\
&radiomics & $0.876$ & $0.881$ & $0.883$ \\
&biomarkers & $0.886$ & $0.895$ & $0.895$ \\
&all & $0.886$ & $0.882$ & $0.895$ \\
&cnn+rad & $0.871$ & $0.894$ & $0.888$ \\
&bio+rad & $0.896$ & $0.893$ & $0.896$ \\
&bio+cnn & $0.874$ & $0.883$ & $0.889$ \\
\bottomrule
\end{tabular}
\label{full_evaluation}
\end{table}

\end{document}